\pdfoutput=1

\documentclass[11pt]{article}

\usepackage[]{EMNLP2023}

\usepackage{float}
\usepackage{url}
\usepackage{graphicx}
\usepackage{booktabs}
\usepackage{multirow} 
\usepackage{footnote}
\usepackage{arydshln}
\usepackage{tablefootnote}
\usepackage{booktabs}  
\usepackage{threeparttable} 
\usepackage{tabularx}

\usepackage{times}
\usepackage{latexsym}

\usepackage[T1]{fontenc}

\usepackage[utf8]{inputenc}

\usepackage{microtype}

\usepackage{inconsolata}

\title{Meta-learning For Vision-and-language Cross-lingual Transfer}

\author{Hanxu Hu\\
  University of Edinburgh \\
  School of Informatics \\
  \texttt{hanxu.hu@ed.ac.uk} \\\And
  Frank Keller \\
  University of Edinburgh \\
  School of Informatics \\
  \texttt{keller@inf.ed.ac.uk} \\}

\begin{document}
\maketitle
\begin{abstract}
Current pre-trained vision-language models (PVLMs) achieve excellent performance on a range of multi-modal datasets. Recent work aims at building multilingual versions of such models, and a range of multilingual multi-modal datasets have been introduced for this purpose. However, current PVLMs typically perform poorly on such datasets when used for zero-shot or few-shot cross-lingual transfer, especially for low-resource languages. To alleviate this problem, we propose a novel meta-learning fine-tuning framework. Our framework makes it possible to rapidly adapt PVLMs to new languages by using Model-agnostic Meta-learning (MAML) in a novel cross-lingual multi-modal manner. Experiments show that this new method boosts the performance of current PVLMs in both zero-shot and few-shot settings on four different vision-language tasks across 14 languages.
\end{abstract}

\section{Introduction}

Multi-modal models focus on jointly learning representations from multiple modalities, such as vision and language. Many tasks require the integration of information of vision and language, including image captioning \cite{vinyals2015show}, natural language visual reasoning \cite{zhou2017visual, suhr2019corpus}, and cross-modal retrieval \cite{zhen2019deep}. Multi-modal learning captures the interaction between different modalities, allowing the resulting representations to be used in multimedia applications that enhance human-computer interaction. 

\begin{figure}[t]
\centering
\includegraphics[width=7cm]{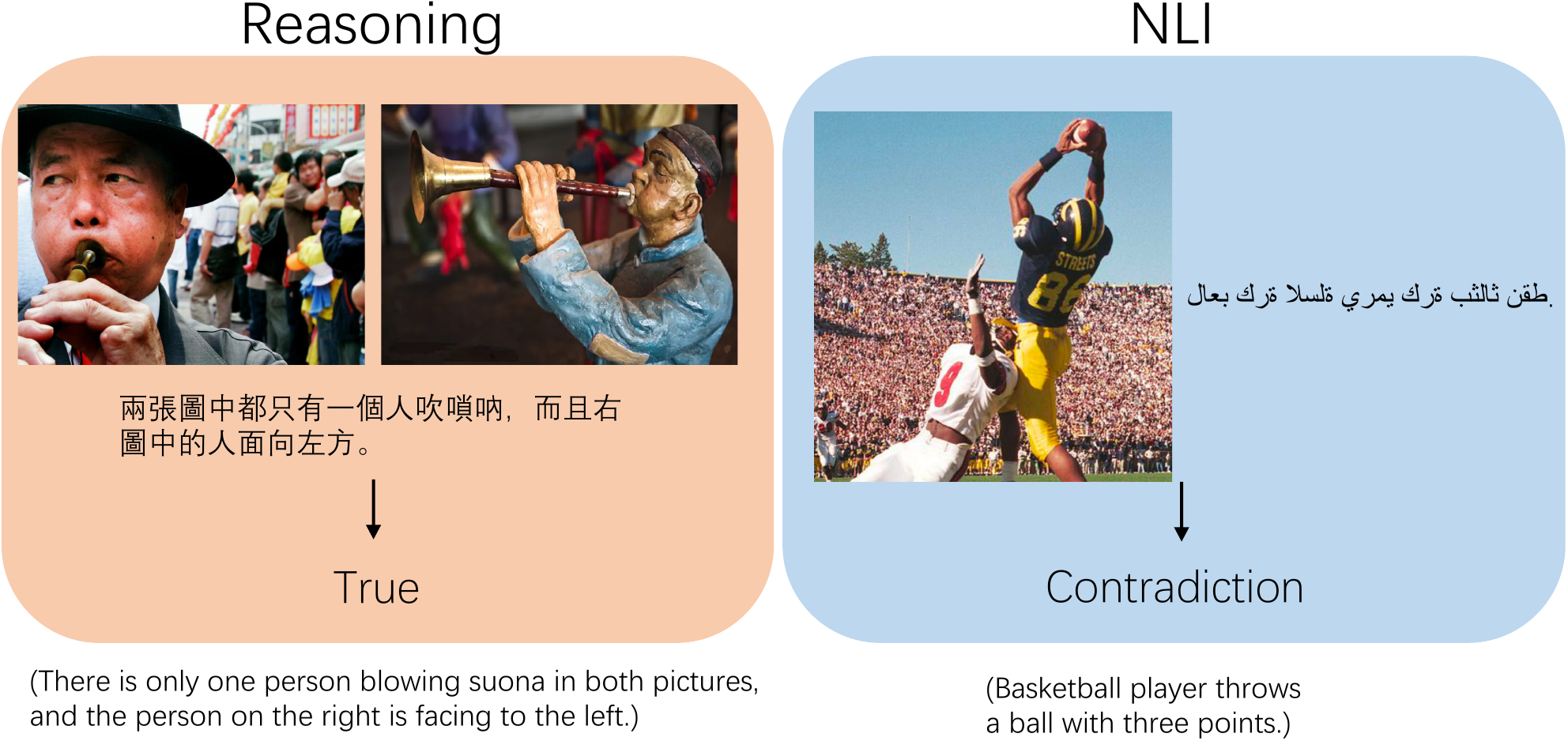}
\caption{Examples in IGLUE \cite{bugliarello2022iglue} benchmark. The left example comes from MaRVL \cite{liu2021visually} dataset, and the right example comes from XVNLI dataset proposed in IGLUE.}
\label{fig:iglueexamples}
\end{figure}

Recently, pre-trained vision-language models (PVLMs; \citealt{chen2020uniter, lu2019vilbert, tan2019lxmert}) have achieved significant advances in multi-modal tasks. However, the data which PVLMs learn from is mostly for high-resource languages such as English. The resulting models rely on large amounts of training data for good performance, and often the models acquire biases that mean they perform poorly in low-resource languages such as Indonesian or Swahili. To address this, several multilingual PVLMs have been proposed \cite{zhou2021uc2, ni2021m3p}. A number of studies have used multilingual multi-modal datasets \cite{bugliarello2022iglue, liu2021visually} and Figure~\ref{fig:iglueexamples} shows two examples from such datasets. The authors of these datasets used them to evaluate current famous PVLMs and demonstrated they do not perform well in low-resource cross-lingual transfer settings.

In this paper, we conjecture that meta-learning can mitigate this issue. This is a learning approach that enables machine learning models to adapt quickly to new tasks by learning the learning algorithm itself. Model-agnostic Meta-learning (MAML; \citealt{finn2017model}) is one of the most widely used meta-learning frameworks. It is based on gradient-descent optimization, does not require multiple models or complex settings, and can be used for a range of models. In previous work \cite{verma2020meta, finn2017model, nooralahzadeh2020zero}, MAML-based methods have been shown to be useful in low-resource and cross-lingual transfer scenarios, including both few-shot and zero-shot cross-lingual tasks. However, prior work has only attempted to use MAML for cross-lingual transfer in \textbf{text-only tasks} \cite{nooralahzadeh2020zero}.

Inspired by previous works about using MAML for natural language tasks, this paper proposes XVL-MAML, a novel variant of MAML that addresses the limitations of previous PVLMs in \textbf{vision-language tasks} for low-resource cross-lingual transfer. 
Our framework combines a traditional supervised loss for learning down-stream tasks with a contrastive loss to encourage the alignment between modalities, resulting in a cross-lingual, multi-modal MAML optimization procedure. 

The intuition underlying our method is that a contrastive loss can align representations of different modalities, and MAML allows the model to generalize quickly to unseen tasks (languages, in our case). We show that XVL-MAML can lead to significant improvements in PVLM performance for cross-lingual transfer. We also find that using contrastive learning in a MAML framework on its own can bring improvements in PVLM performance in unsupervised settings.


In sum, our contributions are as follows: (1)~We propose a novel framework called XVL-MAML which is the first meta-learning method specialized for vision-language cross-lingual transfer, and doesn't require the translation or pre-training data. (2)~We show that using only contrastive learning in the MAML framework in an unsupervised setting can also be useful. (3)~We demonstrate that our proposed framework can boost the performance of current PVLMs across 14 languages and four tasks in both \textbf{zero-shot learning} and \textbf{few-shot learning}.  (4)~We conduct an ablation study to verify the effect of contrastive learning in both supervised and unsupervised settings and present an analysis across languages and tasks.

\section{Related Work}

\subsection{Multilingual Vision-and-Language Methods and Tasks}

Recent work has investigated vision-and-language cross-lingual transfer tasks. \citet{elliott2016multi30k} proposed Multi30K, an image description dataset which contains descriptions in multiple languages. Previous methods \cite{gella-etal-2017-image, rotman-etal-2018-bridging} propose ways of bridging languages through images, but they mainly focus on image-text retrieval and only consider high-resource languages such as English and German. \citet{pfeiffer2022xgqa} built a multilingual visual question answering dataset xGQA. \citet{liu2021visually} proposed a multilingual version of the grounded visual reasoning dataset MaRVL, which follow the same setting as the natural language visual reasoning dataset NLVR2 \cite{su2019vl}, but considers both cross-lingual transfer and domain shift between languages. 

Several pre-trained models are recently proposed for vision-and-language cross-lingual transfer. \citet{ni2021m3p} proposed M3P, a transformer-based pre-trained model that maps the same concepts in different modalities and languages into a common semantic space. Similar to M3P, \citet{liu2021visually} extended UNITER \cite{chen2020uniter}, proposing mUNITER based on M-BERT \cite{devlin2019bert}, and xUNITER based on XLM-R \cite{conneau2020unsupervised}. \citet{zhou2021uc2} proposed UC2, a model using a data augmentation method based on machine translation for cross-lingual cross-modal pre-training. Although pre-training methods have proven powerful across multiple tasks, they require large amounts of training data and show a clear performance gap between English and low-resource languages on the IGLUE benchmark \cite{bugliarello2022iglue}. 

Recently, some adapter-based efficient tuning methods \cite{pfeiffer2022xgqa, wang2023adapting} and translation augmented methods \cite{qiu2022multilingual} were proposed for multilingual multimodal tasks. But they still require a large amount of data or machine translated data for training. Our method, in contrast, only requires a small amount of auxiliary data.  

\subsection{Meta-Learning}
Meta-learning has been increasingly popular in machine learning. Whereas conventional machine learning methods learn by data points, meta-learning learns by tasks. Previous meta-learning work \cite{vinyals2016matching,finn2017model} focused on adapting to new tasks quickly. But meta-learning can be applied to other scenarios as well, including semi-supervised learning \cite{ren2018meta}, multi-task learning \cite{yu2020gradient}, and domain generalization \cite{li2018learning}. 

Prior work has also explored the effectiveness of meta-learning in NLP: \citet{wang2021meta} applied meta-learning in semantic parsing for domain generalization based on MAML \cite{finn2017model, li2018learning}. \citet{relationclass} leveraged meta-learning under limited supervision in a relation classification task. Recently, there have been some applications using MAML in cross-lingual transfer: \citet{gu-etal-2018-meta} and \citet{nooralahzadeh2020zero} regard languages as tasks in their meta-learning framework. In contrast to these existing approaches, which explore text-only scenarios, we are the first to utilize meta-learning for cross-lingual transfer in multi-modal tasks.

\section{Meta-learning for Vision-and-language Cross-lingual Transfer}
 We first formally define the problem of vision-and-Language cross-lingual transfer in the context of zero-shot and few-shot scenarios in Section~\ref{sec:definition}. Then, we introduce our overall fine-tuning framework in Section~\ref{sec:overall}. And we introduce the contrastive learning used for vision-and-language tasks in Section~\ref{sec:contrastive}. Finally, we introduce our XVL-MAML algorithm in Section~\ref{sec:xvl-maml}.


\subsection{Problem Definition}
\label{sec:definition}
Following the multilingual vision-language IGLUE benchmark \cite{bugliarello2022iglue}, we formulate the problem of cross-lingual transfer learning in vision-and-language scenarios. 
For understanding tasks, the input is a pair of an image $V$ and text $U$, and the output $Y$ is the result inferred by the multi-modal model. We can thus formulate this problem as computing $P_\theta(Y | V, U )$, where $\theta$ are the parameters of the PVLMs. During training, the image-text pairs come from datasets $D_s$ in a set of source languages, and our aim is to perform well on datasets $D_t$ for the same task in the target languages. For the zero-shot setup, the pre-trained model fine-tuned on $D_s$ is directly used in inference on $D_t$ for unseen target languages. For the few-shot setup, after training on $D_s$, the model is continually fine-tuned on several shots of the training set of $D_t$ and then evaluated on the development set of $D_t$.

\subsection{Overall Fine-tuning Framework For Cross-lingual Transfer}
\label{sec:overall}

The pipeline of our proposed meta-learning fine-tuning framework can be divided into three parts: 
\begin{enumerate}
    \item Fine-tune the pre-trained vision-language model on data of the down-stream task \textbf{in English}
    \item Fine-tune the model on data in the \textbf{auxiliary language} (one language other than English) using our proposed XVL-MAML algorithm.
    \item Evaluate the fine-tuned model on data in the \textbf{target languages} (languages other than English and the auxiliary language).
\end{enumerate} 

The traditional cross-lingual transfer learning procedure described in \citet{bugliarello2022iglue} only includes part~1 and~3. In part~3, if the setting is zero-shot, the model is evaluated on data in the target language directly, but if the setting is few-shot, the model continues to be fine-tuned on few-shot data in the target languages and is then evaluated. The difference between our framework and the traditional procedure is the additional fine-tuning step of part~2. We will describe it specifically in Section~\ref{sec:xvl-maml}, but before that, we will introduce contrastive learning for vision-and-language tasks.


\subsection{Contrastive Learning for Vision-and-language Tasks}
\label{sec:contrastive}
The vision-and-language contrastive learning loss proposed by \citet{zhang2020contrastive} has proven effective in medical image scenarios and is used as the pre-training objective function of CLIP \cite{radford2021learning}. It can be regarded as an auxiliary task for representation learning, aiming to enable models to gain better aligned multi-modal representation for downstream tasks. In the contrastive learning scheme, a batch of embeddings of images encoded by the model can be written as $I = \{I_{1}, ..., I_{N}\}$, and a batch of embeddings of texts encoded by the model can be written as $T = \{T_{1}, ..., T_{N}\}$, where $N$ is the size of batch, $(I_{i}, T_{i})$ is an image-text pair. If the paired image-text data describe the same or similar concepts, then we can assume they are \textbf{positive} examples, and non-paired data are \textbf{negative} examples. Then, the embeddings of images and texts are fed into two different linear transformation layers separately, $W_1$ and~$W_{2}$: 
\begin{equation}
U = I \cdot W_{1}^\top
\end{equation}
\begin{equation}
V = T \cdot W_{2}^\top
\end{equation}
Where $U$ and $V$ represent the batch of image-text pairs. Then the cosine similarity of each pair can be computed as $\langle U_i,V_j \rangle = \frac{U^{\top}_{i} V_j}{\left\| U_i \right\| \left\| V_j \right\|}$. The objective is to maximize the similarity of matched image-text pairs and minimize the similarity of others. So the image-text contrastive loss can be formulated as follows:
\begin{equation}
\mathcal{L}^{1}_{i} = -\log \frac{\exp(\langle U_i, V_i \rangle)}{\sum^{N}_{K=1}\exp(\langle U_i, V_k \rangle)}
\end{equation}
Following \citet{zhang2020contrastive}, the contrastive loss should be symmetric for each modality, and the text-image contrastive loss is:
\begin{equation}
\mathcal{L}^{2}_{i} = -\log \frac{\exp(\langle V_i, U_i \rangle)}{\sum^{N}_{K=1}\exp(\langle V_i, U_k \rangle)}
\end{equation}
The final contrastive loss of this batch of paired data is then: 
\begin{equation}
\mathcal{L}_{CL} = \sum_{i=1}^{N} (\mathcal{L}^{1}_{i} + \mathcal{L}^{2}_{i})
\end{equation}
Where $\mathcal{L}_{CL}$ is the overall contrastive loss. When we minimize $\mathcal{L}_{CL}$, we maximize the similarity of image-text pairs which are positive examples.

\subsection{XVL-MAML}
\label{sec:xvl-maml}

Inspired by the effectiveness of MAML for quickly adapting to new tasks,  we propose a novel variant of the MAML algorithm specialized for cross-lingual transfer in vision and language tasks, called XVL-MAML. 
Specifically, we first integrate contrastive learning into the MAML algorithm, making it specialized for the visual-language task of cross-lingual transfer learning. Our intuition is that we can use MAML with a contrastive loss as its learning objective for quickly adapting vision-language alignment to new languages. In this framework, the alignment between image and text in a specific language can be regarded as a task. Inspired by \citet{nooralahzadeh2020zero}, we use the data of one auxiliary language for fine-tuning, but with a contrastive loss as objective function in the MAML algorithm. 

Specifically, we sample a batch of support data $\mathcal{B}_s$ and a batch of query data $\mathcal{B}_q$ in the data in auxiliary language $A$ for each virtual task $\mathcal{T}$. Assuming the parameters of the model are $\theta$ and the contrastive loss on the support data is $\mathcal{L}_{CL} (\theta)_{\mathcal{B}_s}$, then the parameters of the model can be updated by one step of gradient descent:
\begin{equation}
\theta^{'} = \theta - \alpha\nabla_\theta \mathcal{L}_{CL} (\theta)_{\mathcal{B}_s}
\end{equation}
Following the MAML algorithm, our final objective for this task is to minimize $\mathcal{L}_{CL} (\theta^{'})_{\mathcal{B}_q}$ on the query data $\mathcal{B}_q$ using gradient descent:
\begin{equation}
\theta \leftarrow \theta -
\beta\nabla_\theta\mathcal{L}_{CL} (\theta^{'})_{\mathcal{B}_q}
\end{equation}
\begin{equation}
\theta \leftarrow\theta - \beta\nabla_\theta\mathcal{L}_{CL} (\theta - \alpha\nabla_\theta \mathcal{L}_{CL} (\theta)_{\mathcal{B}_s})_{\mathcal{B}_q}
\label{eq8}
\end{equation}
Optimized using this method, pre-trained vision-language models can quickly adapt to new tasks in other languages without using any annotation in the auxiliary language for downstream tasks, so we will refer to this as an \textbf{unsupervised scenario}.

In \textbf{supervised scenarios}, where the downstream tasks labels in the auxiliary language are available, we combine the loss of the downstream task $\mathcal{L}$ with the vision-language contrastive loss $\mathcal{L}_{CL}$ by adding them together. So during fine-tuning, Equation~(\ref{eq8}) is modified to:
\begin{equation}
\theta \leftarrow \theta -
\beta( \nabla_\theta\mathcal{L}(\theta^{''})_{\mathcal{B}_q} + \lambda \nabla_\theta\mathcal{L}_{CL} (\theta^{'})_{\mathcal{B}_q} )
\end{equation}
Where the temporary parameters optimized for one step by the downstream task loss $\mathcal{L}$ on the support set $\mathcal{B}_s$ is $\theta^{''}$, $\beta$ is the meta-learning rate, and $\lambda$ is the scale factor of contrastive learning. By simply adding the gradients of the downstream task and contrastive learning in the meta-update, the model learns downstream tasks and vision-language alignment simultaneously for cross-lingual transfer.

\begin{table*}[]
\small
\centering
\begin{tabular}{@{}c|c|ccccc@{}}
\toprule
                                                &                         &                                             &                        &                         & \multicolumn{2}{c}{xFlickr\&Co}             \\ \cmidrule(l){6-7} 
\multirow{-2}{*}{Method}                        & \multirow{-2}{*}{Model} & \multirow{-2}{*}{XNVLI}                     & \multirow{-2}{*}{xGQA} & \multirow{-2}{*}{MaRVL} & IR                   & TR                   \\ \midrule
\multicolumn{1}{c|}{}                           & mUNITER                 & 53.7                                        & 10.0                   & 53.7                    & 8.1                  & 8.9                  \\
\multicolumn{1}{c|}{}                           & xUNITER                 & 59.0                                        & 20.8                   & 56.0                    & 13.8                 & 12.5                 \\
\multicolumn{1}{c|}{}                           & UC2                     & 62.5                                        & 29.0                   & 56.4                    & 19.7                 & 17.0                 \\
\multicolumn{1}{c|}{\multirow{-4}{*}{Baseline}} & M3P                     & 58.2                                        & 28.2                   & 56.0                    & 12.9                 & 11.9                 \\ \midrule
                                                & xUNITER                 & {\color[HTML]{333333} 63.0 \textbf{(+4.0)}} & 22.5 \textbf{(+1.7)}            & \textbf{59.4 (+4.4)}    & 16.3 \textbf{(+2.5)}          & 14.2 \textbf{(+1.7)}          \\
\multirow{-2}{*}{Ours}                          & UC2                     & \textbf{64.4} (+1.9)                      & \textbf{29.9} (+0.9)  & 57.0 (+0.6)             & \textbf{21.3} (+1.6) & \textbf{18.7 (+1.7)} \\ \bottomrule
\end{tabular}
\caption[table1]{Zero-shot performance (accuracy) of four baseline models only fine-tuned on English data (Baseline) and two models fine-tuned by our meta-learning method (Ours) on four IGLUE datasets \cite{bugliarello2022iglue}.}
    \label{tab:zero}
\end{table*}

\section{Experiments}

In this section, we introduce both the base PVLMs we use for vision-language cross-lingual transfer, as well as the datasets and metrics we use to evaluate our proposed method. Then we describe how the experiments were conducted and discuss the results.

\subsection{Base models}
In this paper, we choose xUNITER \cite{liu2021visually} and UC2 \cite{zhou2021uc2} as our base models, as they use different pre-training
methods. Then we applied XVL-MAML to both models to show that this method works across different models.

\paragraph{xUNITER} is a multilingual version of the UNITER model \cite{chen2020uniter}. It has a similar architecture to UNITER and uses Faster-RCNN \cite{ren2015faster} as a feature extractor for images. 
The image features are pooled and reshaped as vectors with the same dimensions as text embeddings. UNITER has four pre-training methods: Masked Language Modelling (MLM), Masked Region Modelling (MRM), Image-Text Matching (ITM), and Word Region Alignment (WRA). xUNITER, in addition to these pre-training methods, also uses Masked Language Modelling for multilingual data and uses the same text embedder as XLM-R \cite{conneau2020unsupervised}.

\paragraph{UC2} uses a similar model architecture as UNITER, but different pre-training methods. Specifically, UC2 augments pre-training on English data by constructing a multilingual corpus via machine translation and then uses this augmented data for pre-training. It also proposes the Visual Translation Language Modeling (VTLM) pre-training method, which uses the image as a pivot to learn the relationship between parallel texts in two languages and their corresponding images.

\subsection{Datasets and Metrics }
We use datasets for four tasks from the IGLUE benchmark \cite{bugliarello2022iglue}, which includes xGQA \cite{pfeiffer2022xgqa}, MaRVL \cite{liu2021visually}, XVNLI, and xFlickr\&Co \cite{plummer2015flickr30k, lin2014microsoft}. We show examples from MaRVL and XVNLI in Figure~\ref{fig:iglueexamples}. Following the convention in IGLUE, the evaluation metric is accuracy for all tasks except cross-modal retrieval, which uses Recall@1. The task format of these four datasets are described below:
\begin{itemize}
    \item 
    \textbf{MaRVL} is a multicutural vision-language reasoning dataset, following the format of English NLVR2 \cite{suhr2019corpus} which  namely to judge whether a sentence is correct or not for a pair of images. 
    \item 
    \textbf{XVNLI} is a multilingual version of visual natural language inference task, which requires models to predict the relationships between premise and hypothesis based on a given image.
    \item 
    \textbf{xGQA} is a multilingual grounded question answering task based on GQA \cite{hudson2019gqa} and machine translated question-answer pairs.
    \item 
    \textbf{xFlickr\&CO} is a multilingual image-text retrieval dataset collected from Flickr30k \cite{plummer2015flickr30k} and COCO \cite{lin2015microsoft}
\end{itemize}

\subsection{Implementation and Hyperparameters }
We conduct all experiments based on the Visiolinguistic Transformer Architectures framework \href{https://github.com/e-bug/volta}{VOLTA} on four 2080Ti GPUs. We implement the MAML algorithm using the \href{https://github.com/facebookresearch/higher}{Higher} library. We use the AdamW \cite{loshchilov2018decoupled} optimizer to fine-tune all models in PyTorch. 

\begin{table*}[t]
\small
\centering
\begin{tabular}{@{}ccccccc@{}}
\toprule
\textbf{METHOD} &
  \textbf{ZH} &
  \textbf{TA} &
  \textbf{SW} &
  \textbf{TR} &
  \textbf{ID} &
  \multicolumn{1}{c}{\textbf{avg}} \\ \midrule
\multicolumn{7}{c}{\textbf{xUNITER}} \\ \midrule
\multicolumn{1}{c|}{Base} &
  54.34/4.74 &
  55.40/6.55 &
  56.41/7.61 &
  57.53/10.99 &
  56.44/7.79 &
  56.02/7.54 \\
\multicolumn{1}{c|}{Ours ($zh\rightarrow X$)} &
  - &
  59.82/14.10 &
  58.85/9.78 &
  60.93/13.22 &
  61.17/13.48 &
 -   \\
\multicolumn{1}{c|}{Ours (avg)} &
  58.34/9.88 &
  58.49/10.25 &
  59.59/10.33 &
  60.06/12.03 &
  60.35/12.41 &
  59.37/10.98 \\
\multicolumn{1}{c|}{Ours (max)} &
  \textbf{59.75/10.28} &
  \textbf{59.82/14.10} &
  \textbf{60.83/10.14} &
  \textbf{62.20/15.25} &
  \textbf{61.17/13.48} &
  \textbf{60.75/12.65} \\ \midrule
\multicolumn{7}{c}{\textbf{UC2}} \\ \midrule
\multicolumn{1}{c|}{Base} &
  57.81/12.25 &
  \textbf{60.06/11.15} &
  51.81/1.09 &
  55.76/7.46 &
  56.56/8.51 &
  56.40/8.09 \\
\multicolumn{1}{c|}{Ours ($zh\rightarrow X$)} &
  - &
  \multicolumn{1}{l}{58.94/12.13} &
  \multicolumn{1}{l}{53.61/7.57} &
  \multicolumn{1}{l}{55.34/7.99} &
  \multicolumn{1}{l}{56.74/8.03} &
  -
   \\
\multicolumn{1}{c|}{Ours (avg} &
  \multicolumn{1}{l}{58.35/13.44} &
  \multicolumn{1}{l}{58.35/12.71} &
  \multicolumn{1}{l}{53.99/7.93} &
  \multicolumn{1}{l}{56.80/9.61} &
  \multicolumn{1}{l}{56.54/9.41} &
  56.81/10.62 \\
\multicolumn{1}{c|}{Ours (max)} &
  \textbf{59.59/13.04} &
  58.94/12.13 &
  \textbf{54.60/9.11} &
  \textbf{58.13/13.48} &
  \textbf{56.74/12.60} &
  \textbf{57.60/12.07} \\ \bottomrule
\end{tabular}
\caption[table1]{Zero-shot performance (accuracy/consistency) of two baseline models fine-tuned only on English data (Base) and then fine-tuned by our meta-learning method (Ours) on the MaRVL dataset \cite{liu2021visually}, where the definition of consistency following \citet{liu2021visually}. Columns indicate target languages. The avg column gives the average performance across all target languages in this row. $zh\rightarrow X$ means the auxiliary language is Chinese, and the target languages is other low-resource languages $X$. We also show the average and maximum performance across all auxiliary languages for each target language.}
    \label{tab:zero1}
\end{table*}

\paragraph{Fine-tuning on English Data}
Before evaluating models on data in low-resource languages, we firstly fine-tune the pre-trained models on the corresponding English datasets: GQA \cite{hudson2019gqa}, NLVR2 \cite{suhr2019corpus}, SNLI-VE \cite{xie2019visual}, and Flickr30k \cite{plummer2015flickr30k} for xGQA, MaRVL, XVNLI, and xFlickr\&Co, respectively, using the procedure of \citet{bugliarello2022iglue} and \citet{liu2021visually}. We follow the setting in IGLUE \cite{bugliarello2022iglue} and also use the IGLUE hyper-parameters for each task when fine-tuning. We save the parameters of models in each epoch, then pick the best performing model for each task as the initialized parameters $\theta$ for the meta-learning fine-tuning stage.

\paragraph{Fine-tuning with Meta-learning}
For the XVL-MAML algorithm, the size of the support set and the query set is~64. We explore learning rates $5\times10^{-5}$, $1\times10^{-5}$, $5\times10^{-6}$ , $1\times10^{-6}$ for both UC2 and xUNITER, and find the best learning rate is $5\times10^{-6}$ for both the normal fine-tuning stage and the meta-update of MAML. For the inner learning rate of XVL-MAML, we explore learning rates $5\times10^{-6}$, $5\times10^{-5}$, $5\times10^{-4}$ and $5\times10^{-3}$, and find that $5\times10^{-4}$ is the best inner learning rate. 

For the proposed meta-learning framework, we find that models overfit after 300 iterations in most situations (for each iterations, we sample a batch of data as support set and a batch as query set), so we set the number of iterations to 400 for all our experiments, and evaluate the performance of models for each 25 iterations to guarantee that we can pick the model with best performance of each setting for evaluation.

\section{Results and Discussion}

\subsection{Zero-shot}

\begin{figure*}[t]
\centering
\hspace{7ex}\includegraphics[width=15cm]{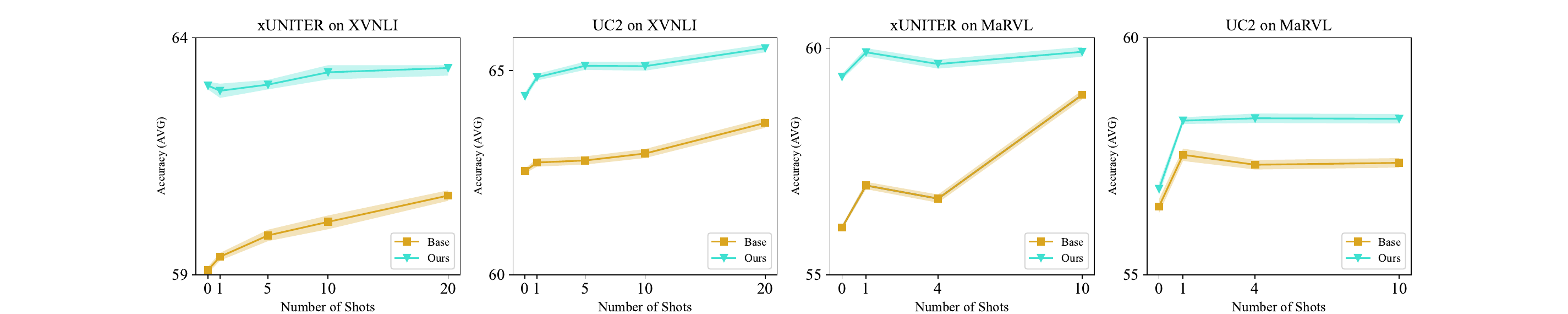}
\caption{Average few-shot performance (accuracy) across all languages of two baseline models on the XVNLI and MaRVL datasets. The horizontal axis represents the number of shots in the training data.}
\label{fig:few_shot}
\end{figure*}

\begin{figure*}[t]
\centering
\hspace{0ex}\includegraphics[width=15cm]{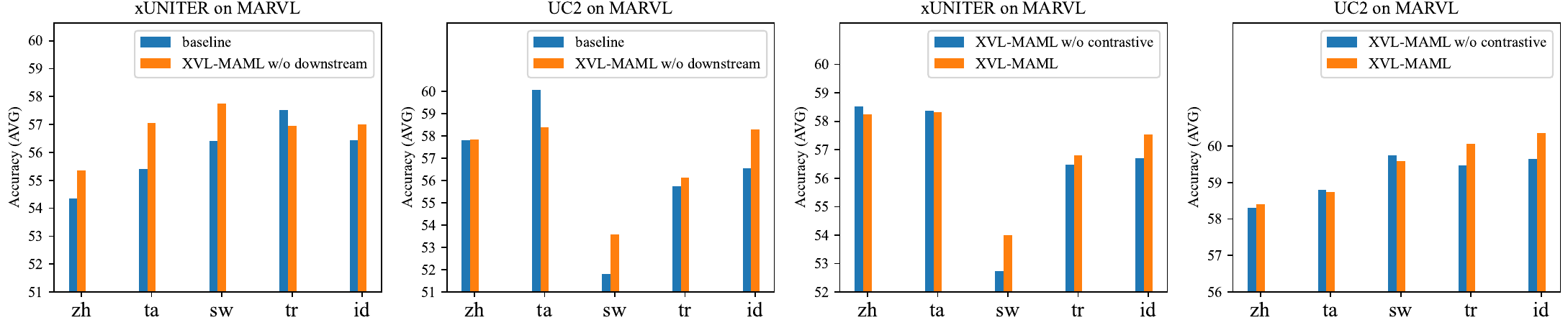}
\caption{Performance in each target languages averaged across auxiliary languages on the MaRVL dataset.}
\label{fig:ablation_plot}
\end{figure*}

We report the results of the baseline models and the results for fine-tuning them using our meta-learning framework in Table~\ref{tab:zero}. In our setting, baseline model means that the PVLM is only fine-tuned on the English datasets. For simplicity, we report the averaged results of all combinations of target languages and auxiliary languages for each model and task. We set the value of $\lambda$ in Equation~(\ref{eq8}) to $2\times 10^{-2}$ for xUNITER and $5\times 10^{-2}$ for UC2 to gain the best performance.

\begin{figure*}[]
\centering
\includegraphics[width=9cm]{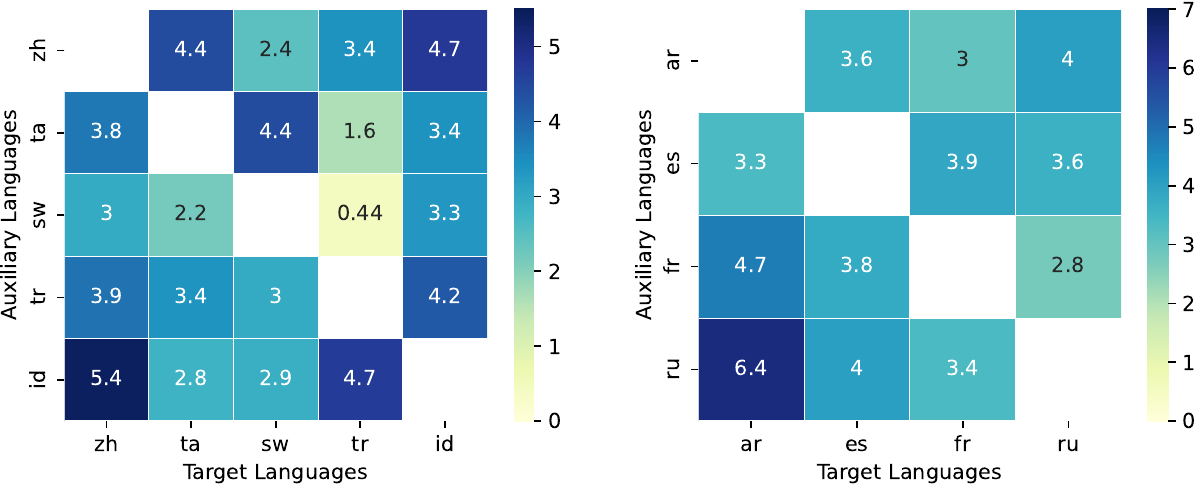}
\caption{Improvements of zero-shot performance by fine-tuning xUNITER on different auxiliary languages then evaluating on different target languages using our proposed framework compared with baseline. The left heatmap is for MaRVL, and the right is for XVNLI. Rows correspond to
auxiliary and columns correspond to target languages.}
\label{fig:heat_map}
\end{figure*}

The results in the Table~\ref{tab:zero} indicate the effectiveness of our meta-learning framework and show that our method can boost the zero-shot performance of UC2 and XUNITER on all four datasets in IGLUE. Note that Table~\ref{tab:zero} shows average performance across all languages. The performance for individual languages can vary, and is shown in detail in Appendix~\ref{sec:appendix}, Table~\ref{tab:all_res}. We also show the differences in improvements when using different auxiliary languages for different target languages in Figure~\ref{fig:heat_map}.

\begin{table}[]
\small
\begin{tabular}{@{}ccc@{}}
\toprule
\multicolumn{3}{c}{Unsupervised Setting}                                               \\ \midrule
\multicolumn{1}{c|}{Method/Models}             & UC2               & xUNITER           \\ \midrule
\multicolumn{1}{c|}{Baseline}                  & 62.5\tiny{$\pm $0.1}           & 59.1\tiny{$\pm $0.1}          \\
\multicolumn{1}{c|}{XVL-MAML(w/o down-stream)}    & \textbf{63.1\tiny{$\pm $0.1} } & \textbf{60.8\tiny{$\pm $0.1} } \\ \midrule
\multicolumn{3}{c}{Supervised Setting}                                                 \\ \midrule
\multicolumn{1}{c|}{Method/Models}             & UC2               & xUNITER           \\ \midrule
\multicolumn{1}{c|}{XVL-MAML(w/o contrastive)} & 63.8\tiny{$\pm $0.1}          & 61.6\tiny{$\pm $0.1}           \\
\multicolumn{1}{c|}{XVL-MAML}      & \textbf{64.4\tiny{$\pm $0.1} } & \textbf{62.9\tiny{$\pm $0.1} } \\ \bottomrule
\end{tabular}
\caption[table2]{Ablation study in the unsupervised setting and supervised setting. The labels of the down-stream task data in the auxiliary language are not given in unsupervised setting and provided in supervised setting.}    \label{tab:un-sup}
\end{table}



\subsection{Few-shot}

\begin{figure*}[t]
\centering
\includegraphics[width=15cm]{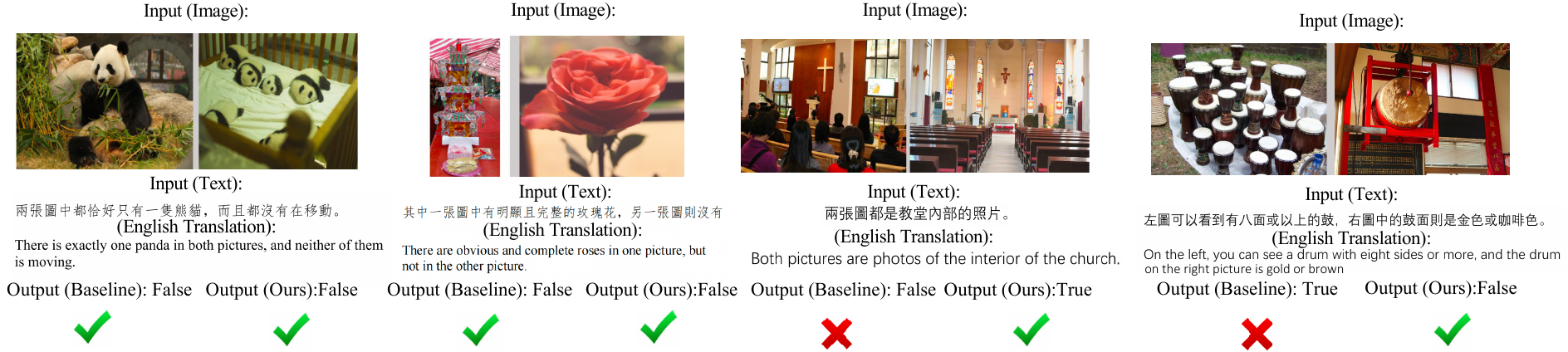}
\caption{Examples from the Chinese part of the MaRVL dataset and predictions of the baseline and ours method.}
\label{fig:case_study}
\end{figure*}

We also conduct few-shot experiments following the setting in IGLUE \cite{bugliarello2022iglue} for both xUNITER and UC2 on XVNLI and MaRVL. The results are shown in Figure~\ref{fig:few_shot}, where the horizontal axis represents the number of shots, and the vertical axis represents the accuracy score. The leftmost point of the horizontal axis is zero, which represents the performance in the zero-shot setup. The blue points and lines show the performance of our method. The yellow points and lines represent the performance of the baseline. We have performed five runs and the interval represents the standard error. It is clear that in all four figures, our method achieves better performance across all shots. And it is worth noting that although there is a slight increase from the performance of zero-shot to one-shot, our proposed method, without seeing any data in the target languages, outperforms the baselines in the few-shot setting, except for UC2 on MaRVL. In other words, only a few instances of training data in target languages are not enough to eliminate the advantage of our method. This demonstrates that while our method requires training data in one auxiliary language, there is no need for few-shot data in the target languages.

\subsection{Ablation Study and Further Analysis}
In this section, we conduct a series of ablation studies which investigate the effect of each part of our proposed meta-learning framework. We have performed five runs for each setting and reported the average and standard error to estimate significant differences.

\paragraph{The Effect of Contrastive Learning} 

We investigate the effect of contrastive learning in our meta-learning fine-tuning framework. Specifically, we fine-tune the model only using a contrastive learning loss in the MAML algorithm (called as "XVL-MAML (w/o down-stream)" in Table~\ref{tab:un-sup}), where the labels of down-stream task data are not given. We evaluate the performance of UC2 and xUNITER on the XVNLI dataset in this setting and reported them in unsupervised setting part of Table~\ref{tab:un-sup}. The results indicate that using contrastive learning solely in the MAML algorithm can improve performance. It provides evidence for the hypothesis that contrastive learning can enable models to learn alignments of modalities in cross-lingual transfer, resulting in better representations.

We also compare the performance of the model in the supervised setting where labels of data in auxiliary language are available; hence in the XVL-MAML algorithm, both contrastive loss and down-stream task loss are used. Then we remove the contrastive learning loss in XVL-MAML, only keeping the down-stream task loss. We compare the performance of these two settings in Table~\ref{tab:un-sup} to show the effectiveness of the contrastive learning loss in XVL-MAML in the supervised setting. In the "Supervised Setting" part of Table~\ref{tab:un-sup}, the first row is XVL-MAML without contrastive learning loss, which means only using down-stream task loss when fine-tuning, and the second row is normal XVL-MAML using both contrastive loss and down-stream task loss. 

Moreover, we show the difference in performance in each target language separately in Figure~\ref{fig:ablation_plot}. Contrastive learning can bring improvements for most of the target languages, especially those whose performance is relatively low when not using contrastive learning. For example, in the leftmost plot, performance in \textit{zh}, \textit{ta}, and \textit{sw} is relatively lower than \textit{tr} in the baseline, but gains significant improvements when using our method. The similar effect can be seen in other three plots and Table~\ref{tab:zero1}.


\paragraph{Diverse down-stream tasks}
We report the results of experiments using four diverse multilingual vision-and-language understanding tasks in Table~\ref{tab:zero}. Our method can bring clear improvements across all tasks for both UC2 and xUNITER, indicating that the approach generalises across tasks. Furthermore, these four IGLUE tasks also differ in the distribution of language families and domains, which indicates our method can be useful across language families and domains. Moreover, our method can significantly boost the performance of xUNITER even in the challenging MaRVL dataset which encompasses five diverse language families and cultures, improving accuracy by 4.4 points.

\paragraph{Diverse languages} 
We also investigate the difference of performance between languages.
Specifically, we take the MaRVL dataset as an example and report results in Table~\ref{tab:zero1}, which lists the performance when using Chinese (zh) as the auxiliary language for meta-learning, and the average and maximum performance across all auxiliary languages for each target language respectively. In most situations, our method results in clear improvements. We then visualize the improvements of xUNITER when using different auxiliary languages for different target languages on MaRVL and XVNLI in Figure~\ref{fig:heat_map}. The improvements we see for MaRVL (which range from 0.44 to 5.4) are smaller than for XVNLI (which range from 2.8 to 6.4), and one possible reason is that the language families of MaRVL are more diverse than those of XVNLI.
But in general, our method improves performance for all combinations of auxiliary and target languages, even when they come from different language families. This further indicates that our method is language-agnostic.

\subsection{Example Predictions}
We show some examples of inputs and predictions for baseline and our method  
in Figure~\ref{fig:case_study}. We use xUNITER to predict the Chinese part of the MaRVL dataset. We have selected two examples where the baseline prediction is incorrect, but our method predicts correctly (the rightmost two examples), and two examples where both our method and baseline method predict correctly (the leftmost two examples). In the two rightmost examples, the label is "True", but the baseline predicts "False". We find that in these two examples, the same concepts ("church" and "drum") described in related texts have different visual features, which makes it more difficult for models to identify them. 
In the left two examples, however, the concepts (panda and roses) described in the text do not have diverse or obscure visual features when they appear in the images. Therefore, based on these cases, we can surmise that the meta-learning framework makes the model more adaptive to diverse information and resulting in better generalization capabilities when mapping between texts and images.

\section{Conclusions}

In this paper, we focused on mitigating the problem of poor performance of current PVLMs in vision-language cross-lingual transfer. We proposed a novel MAML framework to adapt pre-trained models for new languages in vision-and-language tasks. Our framework combines contrastive learning and downstream task supervised learning. We verify the effectiveness of our approach in both supervised and unsupervised settings. The key strength of our method is that we leverage contrastive learning in the MAML procedure so that models can quickly learn to align representations from different modalities and adapt them to unseen languages.

Experimental results demonstrate that our proposed meta-learning framework significantly improves the performance of models in vision-and-language cross-lingual transfer both in zero-shot and few-shot setups. We applied our method to two representative PVLMs, UC2 and xUNITER, and verified its effectiveness on four datasets in the IGLUE benchmark in 14 languages. We also conducted an ablation study to explore the effect of contrastive learning, and analysed the effect of different languages and tasks.

\section*{Limitations}
Our proposed method applies contrastive learning to samples of image-text pairs. The alignments induced in this fashion work best if there is a concept or an object that is both depicted in the image and referred to in the sentence. If this is not the case, then the method may end up learning incorrect alignments; this includes cases where the image or the sentence contain multiple objects or concepts, not all of which can be aligned. To address this limitation, future work should explore how to construct better positive and negative samples and how to enable learning at a more fine-grained level. Besides, current famous PVLMs are encoder-only models, which is different with recent decoder-only LLMs, so meta-learning methods for multi-modal multilingual LLMs is worth to explore as a future work.

\section*{Ethics Statement}

The use of the IGLUE benchmark in our paper is consistent with its intended use. We have checked the datasets for offensive content by sampling and visualizing examples. There are 14 languages in the datasets we use, we list them in Table~\ref{tab:all_res}. More detailed information about the IGLUE dataset can be found in \cite{bugliarello2022iglue}.

\bibliography{anthology,custom,custom_1}
\bibliographystyle{acl_natbib}

\appendix

\section{Appendix}
\label{sec:appendix}



\begin{table*}[tb]
\resizebox{\textwidth}{!}{%
\begin{tabular}{@{}ccccccccccccccc@{}}
\toprule
                                       & Ar             & Bn             & De             & Es             & Id             & Fr             & Ja             & Ko             & Pt             & Ru             & Sw             & Ta             & Tr             & Zh             \\ \midrule
\multicolumn{15}{c}{MaRVL}                                                                                                                                                                                                                                                           \\ \midrule
\multicolumn{1}{c|}{xUNITER (Baseline)} & -              & -              & -              & -              & 56.44          & -              & -              & -              & -              & -              & 56.41          & 55.40          & 57.53          & 54.34          \\
\multicolumn{1}{c|}{UC2 (Baseline)}     & -              & -              & -              & -              & 56.56          & -              & -              & -              & -              & -              & 51.81          & \textbf{60.06} & 55.76          & 57.81          \\
\multicolumn{1}{c|}{xUNITER (Ours)}     & -              & -              & -              & -              & \textbf{60.35} & -              & -              & -              & -              & -              & \textbf{59.59} & 58.49          & \textbf{60.06} & \textbf{59.75} \\
\multicolumn{1}{c|}{UC2 (Ours)}         & -              & -              & -              & -              & 56.74          & -              & -              & -              & -              & -              & 54.60          & 58.94          & 58.13          & 59.59          \\ \midrule
\multicolumn{15}{c}{XVNLI}                                                                                                                                                                                                                                                           \\ \midrule
\multicolumn{1}{c|}{xUNITER (Baseline)} & 53.52          & -              & -              & 60.05          & -              & 61.60          & -              & -              & -              & 61.25          & -              & -              & -              & -              \\
\multicolumn{1}{c|}{UC2 (Baseline)}     & 58.36          & -              & -              & \textbf{63.86} & -              & 65.01          & -              & -              & -              & 64.72          & -              & -              & -              & -              \\
\multicolumn{1}{c|}{xUNITER (Ours)}     & 56.70          & -              & -              & 60.91          & -              & 68.64          & -              & -              & -              & 63.91          & -              & -              & -              & -              \\
\multicolumn{1}{c|}{UC2 (Ours)}         & \textbf{59.94} & -              & -              & 62.97          & -              & \textbf{69.41} & -              & -              & -              & \textbf{65,18} & -              & -              & -              & -              \\ \midrule
\multicolumn{15}{c}{xGQA}                                                                                                                                                                                                                                                            \\ \midrule
\multicolumn{1}{c|}{xUNITER (Baseline)} & -              & 11.41          & 33.21          & -              & 32.38          & -              & -              & 13.28          & 20.51          & 17.84          & -              & -              & -              & 17.20          \\
\multicolumn{1}{c|}{UC2 (Baseline)}     & -              & 19.49          & 33.52          & -              & 29.83          & -              & -              & 23.29          & 31.23          & 32.61          & -              & -              & -              & 33.25          \\
\multicolumn{1}{c|}{xUNITER (Ours)}     & -              & 12.46          & 34.10          & -              & \textbf{33.63} & -              & -              & 15.05          & 22.71          & 20.27          & -              & -              & -              & 19.27          \\
\multicolumn{1}{c|}{UC2 (Ours)}         & -              & \textbf{19.63} & \textbf{34.50} & -              & 29.58          & -              & -              & \textbf{24.93} & \textbf{32.47} & \textbf{33.24} & -              & -              & -              & \textbf{35.00} \\ \midrule
\multicolumn{15}{c}{Xflickr\&Co (IR)}                                                                                                                                                                                                                                                 \\ \midrule
\multicolumn{1}{c|}{xUNITER (Baseline)} & -              & -              & 14.70          & 16.40          & 15.15          & -              & 9.55           & -              & -              & 14.75          & -              & -              & 8.85           & 17.20          \\
\multicolumn{1}{c|}{UC2 (Baseline)}     & -              & -              & 28.10          & 14.65          & 13.55          & -              & 23.70          & -              & -              & 18.20          & -              & -              & 8.15           & 31.70          \\
\multicolumn{1}{c|}{xUNITER (Ours)}     & -              & -              & 16.20          & \textbf{18.85}         & \textbf{18.50} & -              & 12.10          & -              & -              & 17.75          & -              & -              & \textbf{11.10} & 19.40          \\
\multicolumn{1}{c|}{UC2 (Ours)}         & -              & -              & \textbf{29.35} & 16.90 & 14.25          & -              & \textbf{25.15} & -              & -              & \textbf{20.50} & -              & -              & 10.50          & \textbf{32.10} \\ \midrule
\multicolumn{15}{c}{Xflickr\&Co (TR)}                                                                                                                                                                                                                                                 \\ \midrule
\multicolumn{1}{c|}{xUNITER (Baseline)} & -              & -              & 14.2           & 15.45          & 13.95          & -              & 8.30           & -              & -              & 13.15          & -              & -              & 7.75           & 14.4           \\
\multicolumn{1}{c|}{UC2 (Baseline)}     & -              & -              & 23.55          & 11.90          & 10.35          & -              & 22.75          & -              & -              & 17.50          & -              & -              & 6.15           & 26.85          \\
\multicolumn{1}{c|}{xUNITER (Ours)}     & -              & -              & 15.50          & \textbf{16.15}         & \textbf{16.70} & -              & 9.90           & -              & -              & 15.70          & -              & -              & \textbf{9.50}  & 15.75          \\
\multicolumn{1}{c|}{UC2 (Ours)}         & -              & -              & \textbf{25.30} & 13.95 & 12.45          & -              & \textbf{23.50} & -              & -              & \textbf{19.80} & -              & -              & 8.30           & \textbf{27.45} \\ \bottomrule
\end{tabular}}
\caption[table]{Accuracy scores for each target language individually averaged over auxiliary languages.}
\label{tab:all_res}
\end{table*}

This is a section in the appendix.

\end{document}